\documentclass{article}
\usepackage[preprint, nonatbib]{neurips_2024}
\usepackage[dvipsnames,table]{xcolor}
\usepackage{float}
\usepackage{rotating}
\usepackage{times}
\usepackage{url}
\usepackage{latexsym}
\usepackage{tabularx}
\usepackage{makecell}
\usepackage{etoolbox}          
\usepackage[hidelinks]{hyperref}

\usepackage{multirow}
\usepackage{multicol}
\usepackage{booktabs}
\usepackage{appendix}
\usepackage{dashrule} 
\usepackage{enumitem} 
\usepackage{CJKutf8}
\usepackage[most]{tcolorbox}
\usepackage[hang,flushmargin]{footmisc} 
\usepackage[T1]{fontenc}
\usepackage{lipsum}
\usepackage{fancyhdr}
\usepackage{caption}
\usepackage{listings}
\usepackage{ulem} 

\makeatletter
\providecommand{\HyOrg@maketitle}{}

\providecommand{\HyOrg@select@alternate}[2]{#1}
\makeatother

\title{SARI: Structured Audio Reasoning via Curriculum-Guided Reinforcement Learning}
\author{Cheng Wen\thanks{Equal contribution}, Tingwei Guo\footnotemark[1], Shuaijiang Zhao ,Wei Zou\thanks{Corresponding author}, Xiangang Li  \\
\textbf{}\\
Beike Inc., Beijing, China  \\
\texttt{\{wencheng008,zouwei026,lixiangang002\}@ke.com}}

\begin{document}
\maketitle

\begin{abstract}
Recent work shows that reinforcement learning (RL) can markedly sharpen the reasoning ability of large language models (LLMs) by prompting them to "think before answering". Yet whether and how these gains transfer to audio‑language reasoning remains largely unexplored. We extend the Group‑Relative Policy Optimization (GRPO) framework from DeepSeek‑R1 to a Large Audio‑Language Model (LALM), and construct a 32k sample multiple‑choice corpus. Using a two‑stage regimen supervised fine‑tuning on structured and unstructured chains‑of‑thought, followed by curriculum‑guided GRPO, we systematically compare implicit vs. explicit, and structured vs. free‑form reasoning under identical architectures. Our structured audio reasoning model, SARI (Structured Audio Reasoning via Curriculum-Guided Reinforcement Learning), achieves a 16.35\% improvement in average accuracy over the base model Qwen2-Audio-7B-Instruct. Furthermore, the variant built upon Qwen2.5-Omni reaches state-of-the-art performance of 67.08\% on the MMAU test-mini benchmark. Ablation experiments show that on the base model we use: (i) SFT warm‑up is important for stable RL training, (ii) structured chains yield more robust generalization than unstructured ones, and (iii) easy‑to‑hard curricula accelerate convergence and improve final performance. These findings demonstrate that explicit, structured reasoning and curriculum learning substantially enhances audio‑language understanding.


\end{abstract}

\begin{figure*}[h]
    \centering
    \includegraphics[width=15cm]{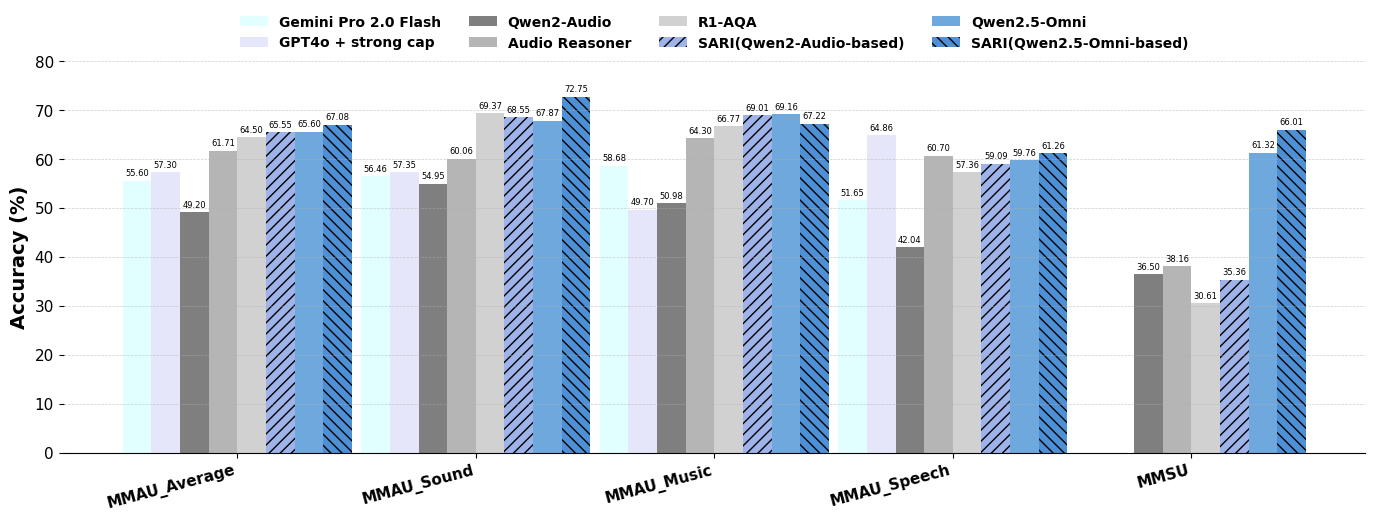}
    \caption{Benchmark Performance of MMAU Test-mini \& MMSU}
    \label{fig:benchmark_mmsu_mmsu}
\end{figure*}

\section{Introduction}
Recent advances in large language models (LLMs) have shown that reasoning capabilities can be greatly enhanced through techniques like chain-of-thought prompting\cite{wei2022chain,chen2025towards}, diverse thinking frameworks\cite{press2022measuring,liang2023encouraging} and reinforcement learning fine-tuning\cite{jaech2024openai,team2025kimi}. In the language modality, models optimized with RL have achieved breakthrough results on complex tasks such as mathematical problem solving and coding challenges. 
These successes suggest that reinforcement learning, particularly using efficient strategies like GRPO\cite{shao2024deepseekmath}, can serve as a general mechanism to boost reasoning in AI systems beyond what supervised learning alone achieves.
In addition, different ways of thinking will also affect model performance. \cite{wen2025thinkpatterns} systematically studied the impact of different thinking patterns on language models of different sizes, which discovered that smaller models could benefit from structured thinking patterns, while larger models perform better with unstructured approaches.

By contrast, audio-language reasoning remains relatively underexplored. Large Audio-Language Models (LALMs) such as Audio Flamingo\cite{kong2024audio}, SALMONN\cite{tang2023salmonn}, and Qwen2-Audio \cite{chu2024qwen2audio} have pushed the state-of-the art in audio understanding. For instance, Qwen2-Audio is capable of following spoken instructions and even outperforms previous systems on audio-centric benchmarks. 
Yet, these models are primarily trained for perception and straightforward QA tasks. They do not incorporate explicit multi-step reasoning or chain-of-thought during training. Audio-Reasoner\cite{xie2025audio} fine-tunes Qwen2-Audio with a structured chain-of-thought approach.  \cite{li2025reinforcement} applied the GRPO algorithm on audio QA tasks. Howover, they observed that simply incorporating an explicit reasoning chain did not yield significant gains in their case, leaving the question of how to best leverage RL training paradigm for audio QA.

In this work, we tackle this challenge by systematically extending reinforcement learning for reasoning to the audio modality. We build upon the GRPO-based RL framework from DeepSeek-R1 and adapt it to train a Large Audio-Language Model to reason about audio. Concretely, we leverage a corpus of approximately 32k multiple-choice audio questions, each with four answer options. Meanwhile, we conduct a comprehensive analysis of different reasoning strategies between explicit thinking and implicit thinking, and between structured thinking and unstructured thinking. Another key aspect of our approach is the introduction of curriculum learning in the RL fine-tuning phase. Rather than exposing the policy model  to the full difficulty spectrum of questions from the start, we rank the training samples by their difficulty level. Curriculum-guided\cite{bengio2009curriculum} RL has the intuitive benefit of preventing the policy from being overwhelmed by very difficult, low-reward episodes early on. This ease-to-difficulty curriculum significantly stabilizes training and improves the model’s reasoning ability. 

In summary, we present a RL-based training regimen for audio-language reasoning, with a curriculum-driven optimization schedule. We show that our approach yields state-of-the-art performance on complex audio QA benchmarks, outperforming conventional supervised fine-tuning as well as prior RL baselines, as showd in Figure \ref{fig:benchmark_mmsu_mmsu}. The contributions of this paper are as follows: 
\begin{itemize}
     \item   \textbf{RL Extension to Audio Reasoning}: We propose extension of reinforcement learning fine-tuning to the audio modality, applying the GRPO algorithm to improve an audio-language model’s reasoning abilities on audio-based tasks.
     \item \textbf{Comparative Study of Reasoning Strategies under RL}: we perform a systematic comparison of multiple reasoning strategies, specifically focusing on explicit vs. implicit reasoning and structured vs. unstructured thinking process. 
     \item \textbf{Improved Performance and Reasoning Quality}: Our RL-trained audio reasoning model achieves new state-of-the-art results on benchmark audio reasoning evaluations (e.g., the MMAU\cite{sakshi2024mmau}), surpassing prior SFT and RL approaches . 
\end{itemize}
In the following, we review related works in Section 2, followed by detailed descriptions of our dataset creation process and training methodology in Section 3. Section 4 presents the experimental setup and evaluation metrics, while Section 5 systematically documents our experimental results, including ablation studies on reasoning styles, cold start strategies, and curriculum learning. We conclude in Section 6 with insights and future directions for advancing audio multi-modal reasoning research.

\section{Related works}
\textbf{Language Reasoning Models}. Recently, models like OpenAI-o1\cite{jaech2024openai}, Kimi K1.5\cite{team2025kimi} and DeepSeek-R1\cite{guo2025deepseek} have attracted widespread attention for their significant performance improvements on reasoning tasks achieved through reinforcement learning\cite{jin2025search,peng2025lmm,openr1}. This paradigm shift has inspired a wave of follow-up research. Some studies have successfully replicated the approach\cite{xie2025logic}, achieving impressive results, while others have focused on refining the algorithm itself to further enhance reasoning efficiency and model performance\cite{yu2025dapo}.

\textbf{Vision Reasoning Language Models}. Reinforcement learning methods for improving reasoning have also started to appear in vision-language\cite{yang2025r1,feng2025video,huang2025vision}. For instance, Vision-R1\cite{huang2025vision} propose Progressive Thinking Suppression Training strategy and employ GRPO to address the post-cold-start overthinking. Video-R1\cite{feng2025video} systematic explored R1 reinforcement learning paradigm for video reasoning in MLLMs. LMM-R1 designed a framework  to enhance multimodal reasoning in LMMs through rule-based RL.

\textbf{Large Audio Language Models}. Large Audio-Language Models (LALMs) like Qwen2-Audio\cite{chu2024qwen2audio}, Audio Flamingo\cite{kong2024audio} and SALMONN\cite{tang2023salmonn} have demonstrated strong audio understanding capabilities. Audio-CoT\cite{ma2025audio} firstly explored the Chain-of-Thought (CoT) reasoning methods in LALM, but they did not involve the update of model parameters and had limited improvement in complex problems. Audio-Reasoner\cite{xie2025audio} introduced a large synthetic dataset (CoTA) with over a million QA pairs and step-by-step reasoning annotations which significantly improved long-context reasoning performance. \cite{li2025reinforcement} applied the GRPO algorithm to fine-tune a Qwen2-Audio 7B model on audio QA tasks . The RL-based approach effectively improved reasoning accuracy despite using far less data, suggesting that reward-driven optimization can extract more signal from limited training examples. 

\section{Methodology}
Our approach consists of three main components: (1) Dataset construction for audio question answering, (2) Reasoning path generation in both structured and unstructured styles using large language models, and (3) Training strategies including supervised fine-tuning and curriculum-based reinforcement learning to teach an audio-language model to reason explicitly. We detail each component below.

\subsection{Data Construction}
\begin{figure*}[h]
    \centering
    \includegraphics[height=6.5cm, width=14cm, trim=0 100 0 80,clip]{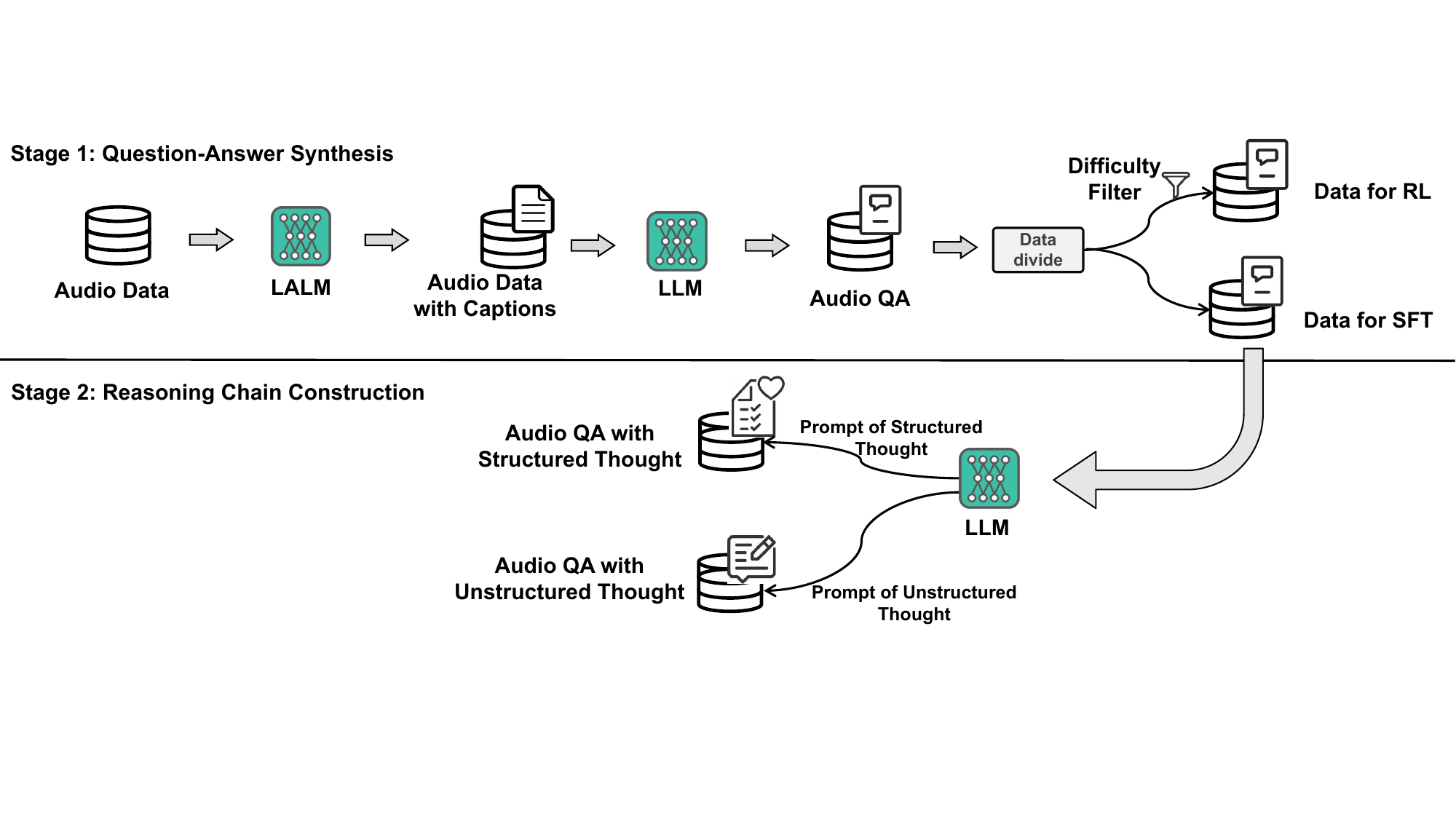}
    \caption{Data Construction}
    \label{fig:data_cons}
\end{figure*}

We built a multiple-choice question-answering dataset focusing on audio understanding and reasoning, and the entire data construction process used open source models and open source datasets, as showed in Figure \ref{fig:data_cons}. The following sections describe each stage in detail.

\textbf{Stage 1}: Question-Answer Synthesis. 
Our data mainly comes from: AudioSet\cite{gemmeke2017audio}, MusicBench\cite{melechovsky2024mustango}, Meld\cite{poria2018meld} and AVQA\cite{yang2022avqa}. We use the Qwen2-Audio-7B-Instruct model to generate an audio caption for each clip in Audioset and Meld, while uses the original caption information in Musicbench. Based on these enriched descriptions, Qwen2.5-72B\cite{yang2024qwen2} is employed to generate a diverse set of questions. At the same time, we randomly selected 20k samples from the AVQA dataset.  In total, we curated approximately 42k multi-choice QA samples. Each sample consists of: an audio input, a question about the audio, four answer options with one correct answer. 

 We divide the dataset into two distinct subsets based on their usage:

\begin{itemize}
    \item \textbf{$Data_{SFT}$ (Supervised Fine-Tuning data)}: This subset is used to supervise the model in generating both reasoning and final answers. It consists of 2,000 samples, each annotated with a reference reasoning path (details provided in the next section). This supervision enables the model to learn fundamental audio concepts and common reasoning patterns.

    \item \textbf{$Data_{RL}$ (Reinforcement Learning data)}: To identify challenging examples, we employ the baseline Qwen2-Audio-7B-Instruct model as a proxy evaluator. For each question in the candidate pool, the model is prompted to generate responses across 16 independent attempts. We compute the \textit{pass rate} as the proportion of trials in which the model provides the correct answer. Questions with a 0\% pass rate—i.e., where the model fails in all 16 attempts—are considered either ambiguous or excessively difficult. These cases are filtered out to avoid introducing unsolvable samples that offer minimal learning signal. After this filtering process, the remaining \textbf{$Data_{RL}$} subset consists of approximately 30,000 samples.
\end{itemize}
\textbf{Stage 2}: Reasoning Chain Construction.
For each question in the supervised set $Data_{SFT}$, we provide a reasoning chain – a series of steps or an explanation leading to the answer. We use the Qwen2.5-72B\cite{yang2024qwen2} model to generate the reasoning path based on the question, answer and the audio caption. We construct two types of reasoning data to supervise the model under different thinking paradigms.  The relevant instruction template can be found in the appendix \ref{AppendixA}.

\begin{itemize}
    \item \textbf{Structured Reasoning (Structured CoT)}: We prompt the model to generate reasoning outputs explicitly divided into four sections: \textit{Planning}, \textit{Caption}, \textit{Reasoning}, and \textit{Summary}. This structured format is inspired by prior works such as \textsc{LLaVA-CoT}\cite{xu2025llavacotletvisionlanguage} and Audio Reasoner\cite{xie2025audio}. In the \textit{Planning} stage, the model analyzes the query to identify key components and outline the reasoning trajectory. The \textit{Caption} stage involves extracting relevant multimodal features from the input, which serve as the grounding for inference. In the \textit{Reasoning} phase, the model performs a step-by-step analysis, integrating audio cues with external knowledge and logical deduction. Finally, the \textit{Summary} section presents the appropriate results based on the reasoning content. We refer to the resulting dataset as $Data_{SFT\_StructuredThought}$.

    \item \textbf{Unstructured Reasoning (Unstructured CoT)}: To model more free-form, naturalistic reasoning, we prompt the model to simulate human-like thought processes from question to answer, without imposing any predefined structure or sectioning. The reasoning is expected to mirror human cognitive behavior—exploring alternative hypotheses, revising assumptions, questioning conclusions, backtracking, validating ideas, and gradually refining judgments. Although unstructured, the model is required to produce a coherent explanation ending with a definitive answer. We denote this dataset as $Data_{SFT\_UnstructuredThought}$.
\end{itemize}
For quality control, multiple candidate reasoning path can be sampled. In our pipeline, we automatically verify that the final answer extracted from the reasoning matches the ground-truth correct option for the question. If not, we regenerate or adjust the prompt until the LLM produces the correct answer with a plausible reasoning chain. This procedure results in high-quality annotated reasoning paths for  \textbf{$Data_{SFT}$} samples.

\begin{figure*}[h]
    \centering
    \includegraphics[width=13cm]{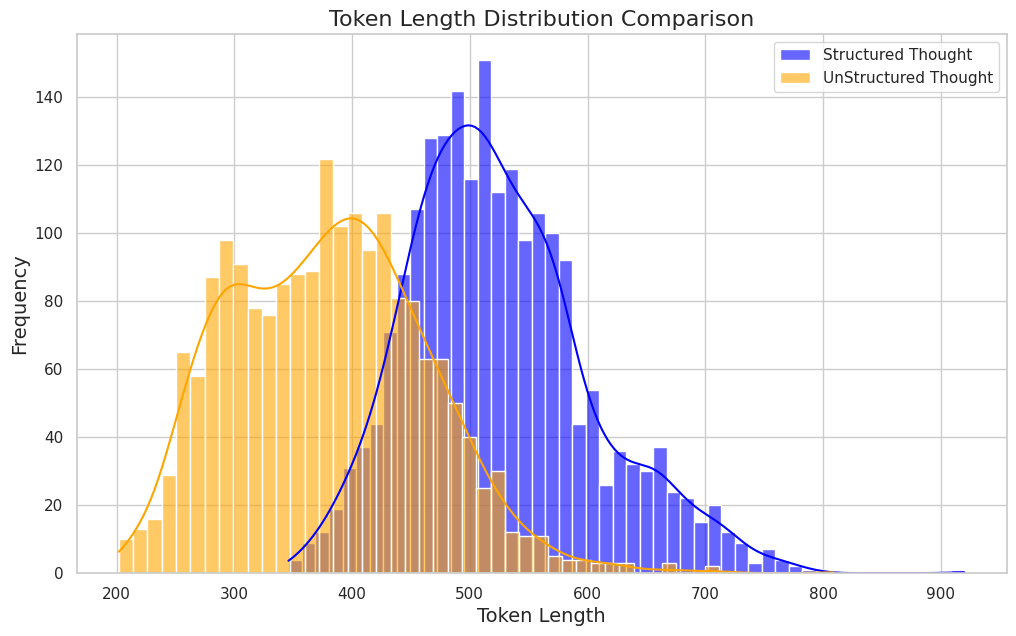}
    \caption{Token length of different Think Pattern}
    \label{fig:text_len}
\end{figure*}

The outcome is a supervised training set containing both question-answer pairs and detailed reasoning. Notably, we have two versions for each question (structured and unstructured reasoning), allowing us to train and compare two different styles of chain-of-thought in our model. We also analyzed the token lengths of $Data_{SFT}$. As illustrated in the \ref{fig:text_len}, structured thinking generally results in longer token sequences compared to unstructured thinking. This difference primarily arises because the structured thought incorporates multiple explicit modules, leading to increased redundancy.

\subsection{Training Strategies for Reasoning Enhancement}

We employ a two-phase training approach for our audio-language model: first supervised fine-tuning (SFT) on the reasoning-augmented data, and then reinforcement learning (RL) fine-tuning  with a reward for correct format and answer. Within this framework, we explore several experimental settings to evaluate the impact of explicit reasoning, the think pattern, cold start and curriculum learning. We detail these strategies below (labeled (a)–(g) corresponding to different training variants):
\begin{itemize}
 \item  \textbf{(a) Implicit Baseline (Direct answer, RL only)}: As a baseline, we fine-tune the model only to output the final answer, without showing any reasoning. We initialize with the pre-trained Qwen2-Audio-7B-Instruct model and directly perform RL on $Data_{RL}$, rewarding the model for choosing the correct option.

 \item  \textbf{(b) Explicit Structured Reasoning, RL from scratch}: In this variant, we attempt to train the model to output a structured chain-of-thought and the answer via RL alone (no supervised warm-up). The model is prompted to produce the four-section reasoning format and then the answer. The reward is entirely dependent on whether the final answer and format are correct.

 \item  \textbf{(c) Explicit Unstructured Reasoning, RL from scratch}: Similar to (b), but the model is prompted to produce an unstructured natural reasoning before the answer. 
\end{itemize}

\begin{tcolorbox}[colframe=blue!60!black, colback=blue!5!white, title=Prompts for GRPO]
\textbf{Implicit Baseline:} Output the final answer in <ANSWER> </ANSWER>

\textbf{Explicit Structured Reasoning:} Output the thinking process in <THINK> </THINK>, and divide it into four parts: <PLANNING>, <CAPTION>, <REASONING>, and <SUMMARY>. After that, output the final answer in <ANSWER> </ANSWER>.

\textbf{Explicit Unstructured Reasoning:} Output the thinking process in <THINK> </THINK> and final answer in <ANSWER> </ANSWER>.

\vspace{0.2cm}
\end{tcolorbox}

The outcomes of (a), (b), (c) informed our next steps: a good initialization is crucial. Simply put, direct RL on $Data_{RL}$ was insufficient to induce correct reasoning or even consistently correct answers – confirming that complex reasoning abilities do not emerge from scratch with sparse rewards. We assume this is mainly due to the weak performance of the Qwen2-Audio-7B-Instruct base.

We therefore employ a two-stage training for the subsequent variants:

    1.    Supervised Fine-Tuning (SFT) Stage: We warm-start the model on $Data_{SFT}$, teaching it to produce the desired output format (either structured or unstructured reasoning + answer). We perform this SFT for each reasoning style separately:
    
    \textbf{Structured SFT:} Train on $Data_{SFT\_StructuredThought}$ examples. We input the question and train the model to output the four-step reasoning followed by the correct answer.
    
    \textbf{Unstructured SFT:} Train on $Data_{SFT\_UnstructuredThought}$ examples, where the output is a free-form rationale ending in the answer.

After this stage, we obtain two candidate models: one that can articulate structured reasoning and one with unstructured reasoning. Both are capable of providing coherent explanations.

    2.    Reinforcement Learning (GRPO) Stage: We then fine-tune each model on $Data_{RL}$ using the GRPO algorithm. GRPO uses a batch of policy outputs and their relative rewards to update the policy without a separate value network, which simplifies training. 
Using this two-stage process, we explore the following setups:
\begin{itemize}
\item \textbf{(d) Structured CoT (SFT + RL)}: We take the structured reasoning model from SFT and fine-tune it with GRPO on $Data_{RL}$. The model’s outputs remain in the Planning/Caption/Reasoning/Summary format. This approach combines explicit structured thinking with outcome-driven refinement. We hypothesize this will yield a model that can tackle difficult questions by systematically breaking them down.

\item \textbf{(e) Unstructured CoT (SFT + RL)}: Likewise, we fine-tune the unstructured reasoning model on $Data_{RL}$ with GRPO. This model will produce more free-form explanations as it optimizes for accuracy. It may have more flexibility in expression, but potentially less consistency in how it reasons compared to the structured one.
\end{itemize}

Within the RL training on $Data_{RL}$, we incorporate a curriculum by ordering the training samples from easier to harder. Concretely, we rank the $Data_{RL}$ questions by the baseline model’s pass rate (from highest to lowest success). This ease-to-difficulty schedule ensures the model first learns to confidently solve questions that are within reach, receiving positive rewards more frequently, and then is gradually challenged with more difficult ones.
We evaluate two curriculum-augmented variants:
\begin{itemize}
\item \textbf{(f) Structured CoT + Curriculum RL}: Different from model (d), we take the structured reasoning model from SFT and  apply curriculum during GRPO training. 

\item \textbf{(g) Unstructured CoT + Curriculum RL}: Similarly, for the unstructured reasoning model (e), we train with a curriculum schedule. We compare this to (e) to quantify the benefit of curriculum learning for a free-form reasoning model.
\end{itemize}

In summary, our training strategies allow us to assess: implicit vs. explicit reasoning (comparing model (a) to models (b)/(c)), structured vs. unstructured reasoning styles (comparing (d) vs. (e) under similar conditions), and impact of cold start (comparing (d) vs. (b), and (e) vs. (c)) and curriculum (comparing (f) vs. (d), and (g) vs. (e)). All models are based on the same underlying architecture (Qwen2-Audio and Qwen2.5-Omni) and are evaluated on the same benchmarks for a fair comparison.

\section{Experimental Setup}

\subsection{Model Setup}
We used Qwen2-Audio-7B-Instruct and Qwen2.5-Omni as the basic models for experiments. Due to the limitation of training resources, we completed all experimental strategies (a)-(g) based on Qwen2-Audio-7B-Instruct, and only partially verified the experiments on Qwen2.5-Omni. During the SFT phase, each model was trained for 3 epochs on the $Data_{SFT}$ dataset with a total batch size of 64 and an initial learning rate of 2e-5. In the GRPO phase, models underwent 1 epoch of training on the $Data_{RL}$ dataset using a total batch size of 32 and an initial learning rate of 1e-6. The KL divergence penalty was disabled by setting its coefficient to 0. The temperature parameter was set to 1.0, with 4 responses generated per sample at each GRPO optimization step. 
\subsection{Evaluation Metrics}
We evaluate model performance primarily by accuracy on multi-choice questions. Two main evaluation sets are used:
\begin{itemize}
    \item   \textbf{MMAU Benchmark}\cite{sakshi2024mmau}: We use the official test-mini set of the MMAU benchmark, which contains challenging audio QA pairs requiring expert-level reasoning. This test set evaluates both in-domain performance on tasks similar to our training data (since we use some AudioSet and music questions). Accuracy is calculated as the percentage of correctly answered questions (multi-choice questions).
    \item   \textbf{MMSU Evaluation}\cite{chen2024voicebench}: To assess generalization, we evaluate on a collection of MMSU questions. MMSU is a diverse multi-choice QA set derived from the MMLU-Pro benchmark (a variant of Massive Multitask Language Understanding) focusing on non-math, non-coding domains. It covers 12 different knowledge areas (history, science, etc.) and was used in the VoiceBench study to test voice assistant knowledge. This evaluates if our audio reasoning training conferred broader reasoning skills (or at least did not hurt them). 
\end{itemize}

For each model, we generated four independent responses and computed the final performance metric as the average accuracy across these trials. Notably, for evaluations on the MMAU test-mini dataset, we utilized Qwen2-72B-Instruct to assess answer correctness by jointly analyzing the question, reference answer, and model outputs, thereby ensuring a reliable and robust evaluation.

\section{Experimental Results and Analysis}
As illustrated in Figure \ref{fig:benchmark_mmsu_mmsu}, our models achieved SOTA performance on MMAU test-mini and MMSU benchmarks. To systematically analyze the impact of different reasoning strategies and training methodologies, we conducted ablation studies using Qwen2-Audio-7B-Instruct and Qwen2.5-Omni as the baseline. Detailed experimental results are tabulated in Table \ref{tab:qwen2-audio-result} and Table \ref{tab:qwen2-omni-result} respectively.

\begin{table*}[h!]
\centering
\setlength{\tabcolsep}{4pt} 
\caption{Ablation Study Employing Qwen2-Audio-7B-Instruct as the Base Model. Decoupling effects of reasoning strategies and training methods, including SFT, GRPO and Curriculum Learning (CL).}
\scalebox{0.72}{
\begin{tabular}{cccccccc}
\toprule
\midrule

\multirow{3}{*}{\textbf{Model}} & \multirow{3}{*}{\textbf{Method}} & \multirow{3}{*}{\textbf{Reasoning Strategy}} & \multicolumn{4}{c}{\textbf{MMAU Test-mini}} & \multicolumn{1}{c}{\textbf{MMSU}} \\
\cmidrule(lr){4-7} \cmidrule(lr){8-8} 
& & & Sound$\uparrow$ & Music$\uparrow$ & Speech$\uparrow$ & Average$\uparrow$ & Accuracy $\uparrow$ \\
\midrule
\midrule
 \multicolumn{1}{l}{\uline{\textbf{\textit{Baselines:}}}} \\
 Qwen2-Audio-7B-Instruct & - & - & 54.95 & 50.98 & 42.04 & 49.20 & 35.72 \\
\midrule
\midrule
 \multicolumn{1}{l}{\uline{\textbf{\textit{Effectiveness of GRPO:}}}} \\
 model-a & GRPO & w/o CoT & 59.84 & 60.11 & 59.76 & 59.90 & 30.45 \\
 model-b & GRPO & Structured CoT & 63.29 & 59.13 & 56.16 & 59.53 & 30.71 \\
 model-c & GRPO & Unstructured CoT & 63.96 & 63.77 & 58.41 & 62.05 & 31.52 \\
\midrule
\midrule
 \multicolumn{1}{l}{\uline{\textbf{\textit{Necessity of Supervised Warm-up:}}}} \\
 sft-a & SFT & Structured CoT & 64.87 & 59.13 & 44.52 & 56.18 & \textbf{37.29} \\
 sft-b & SFT & Unstructured CoT & 59.91 & 59.58 & 43.54 & 54.35 & 35.04 \\
 model-d & SFT + GRPO & Structured CoT & 70.12 & 65.28 & 54.88 & 63.58 & 35.58 \\
 model-e & SFT + GRPO & Unstructured CoT & 67.27 & 63.25 & 60.44 & 63.65 & 34.32 \\
\midrule
\midrule
 \multicolumn{1}{l}{\uline{\textbf{\textit{Impact of Curriculum Learning:}}}} \\
 model-f & SFT + GRPO + CL & Structured CoT & 68.55 & \textbf{69.01} & 59.09 & \textbf{65.55} & 35.36 \\
 model-g & SFT + GRPO + CL & Unstructured CoT & 67.12 & 63.85 & 60.04 & 63.68 & 32.56 \\
\midrule
\bottomrule
\end{tabular}}
\label{tab:qwen2-audio-result}
\end{table*}

\textbf{Effectiveness of GRPO}: Based on the Qwen2-Audio-7B-Instruct, we employed the GRPO method to enhance the model's implicit and explicit reasoning capabilities respectively, resulting in the generation of models-a, b, and c. In the experimental results of model-a, we observe a trade-off between its performance on in-domain and out-of-domain tasks. Compared to Qwen2-Audio-7B-Instruct, its accuracy on MMAU test-mini improved by 10.7\%, validating the effectiveness of the GRPO approach. However, model-a showed a noticeable decline in accuracy on MMSU. We attribute this to the implicit model being narrowly optimized to choose correct answers, potentially memorizing domain-specific patterns. Although we required models-b and c to possess explicit reasoning capabilities, in the absence of guidance, the reasoning processes of these models were often meaningless. Consequently, both models achieved results similar to model-a.

\textbf{Necessity of Supervised Warm-up (Cold Start)}: When comparing model-d/e with model-b/c, the importance of supervised pre-training in enhancing the model's explicit reasoning capabilities becomes evident. Without a cold start, RL training cannot produce a viable reasoning model. The chain-of-thought outputs from model-b/c are mostly meaningless, preventing them from performing effective reasoning and achieving greater performance improvements. In stark contrast, by initializing via SFT on our reasoning datasets, explicit reasoning can be executed as intended. The accuracy of model-d and e on the MMAU test-mini improved by 4.05\% and 1.60\%, respectively, compared to model-b and c. Additionally, due to the generalization provided by explicit reasoning capabilities, the accuracy of them on MMSU is also significantly higher than that of model-b and c.

\textbf{Impact of Curriculum Learning}: Introducing curriculum learning during the reinforcement learning phase can effectively enhance performance. On the MMAU test-mini dataset, models employing a training schedule from easy to difficult (model-f and g) outperformed those using a randomly shuffled training set (model-d and e). Specifically, the accuracy of model-f improved by 1.97\% compared to model-d, surpassing the current state-of-the-art models derived from the Qwen2-Audio-7B-Instruct. The accuracy of model-g was slightly higher than that of model-e. During training, we observed that the curriculum learning approach converged faster and reached a better optimum. Early in the training process, model-f was able to handle moderately difficult problems and achieve stable rewards, whereas the model-d was often overwhelmed by very difficult samples. This improvement is particularly evident in the context of explicit reasoning: chain-of-thought models have a more complex output space, making it crucial to guide them with a curriculum.

\textbf{Structured vs. Unstructured Reasoning}: Unlike the flexible and variable chain-of-thought in the unstructured reasoning paradigm, we divide the structured reasoning process into four parts: Planning, Caption, Reasoning, and Summary. On the MMAU test-mini and MMSU datasets, models with structured thinking capabilities (model-d and f) generally achieved higher accuracy than those with unstructured thinking capabilities (model-e and g). This demonstrates that the structured reasoning approach helps models avoid errors—for example, by explicitly writing down the identified sounds (Caption) and then analyzing them (Reasoning), making it less likely to overlook critical details. Nonetheless, unstructured models remain valuable as they show that the method does not rely on prompts in a specific format—even freely generated chains of thought can enhance reasoning capabilities.

\begin{figure*}[h]
    \centering
    \includegraphics[width=13cm]{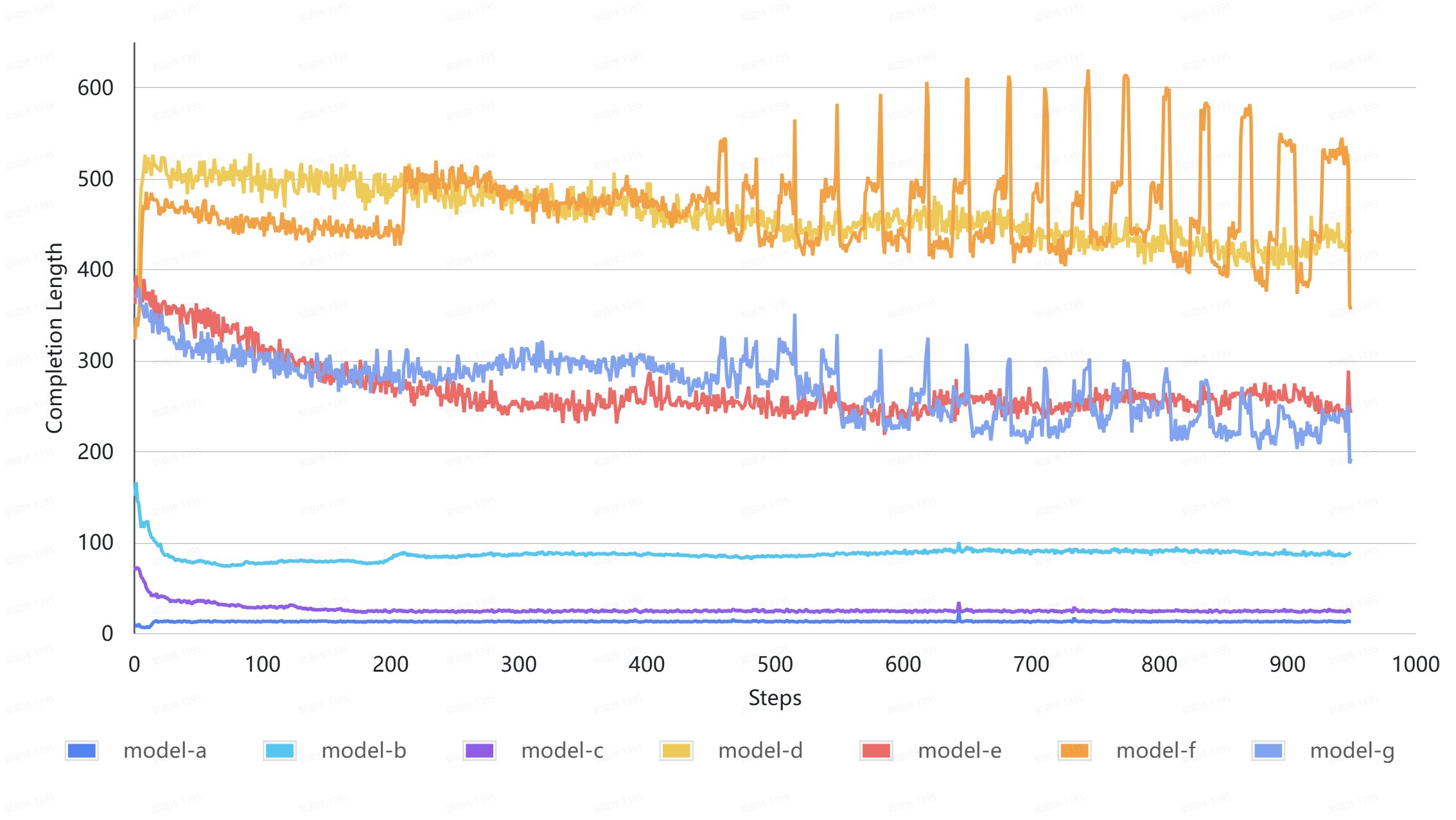}
    \caption{Completion length of our models derived from Qwen2-Audio-7B-Instruct in convergence processes.}
    \label{fig:model}
\end{figure*}

Figure \ref{fig:model} illustrates the variation in completion lengths across different models during the convergence process. Without pretraining via SFT, models fail to develop effective chain-of-thought reasoning, as evidenced by the significantly shorter completion lengths of model-b and c. In addition, models employing structured reasoning paradigms exhibit longer completion lengths compared to those using unstructured approaches, which aligns with the characteristics observed in the training data. Response examples from model-a to g derived from Qwen2-Audio-7B-Instruct can be found in appendix \ref{AppendixB}.

Furthermore, experimental results on Qwen2.5-Omni provide additional validation of these findings. As demonstrated in Table \ref{tab:qwen2-omni-result}, the structured CoT implementation outperformes its unstructured counterpart on the MMAU test-mini benchmark, with accuracy scores of 67.08\% vs.
65.90\%. At the same time, the model performing structured CoT also exhibits significant advantages on MMSU,
attaining a remarkable accuracy of 66.01\%.

\begin{table*}[h!]
\centering
\setlength{\tabcolsep}{4pt} 
\caption{Accuracy Rates (\%) of Our Qwen2.5-Omni-Based Models on MMAU Test-Mini and MMSU Benchmarks}
\scalebox{0.82}{
\begin{tabular}{cccccccc}
\toprule
\midrule

\multirow{3}{*}{\textbf{Model}} & \multirow{3}{*}{\textbf{Method}} & \multirow{3}{*}{\textbf{Reasoning Strategy}} & \multicolumn{4}{c}{\textbf{MMAU Test-mini}} & \multicolumn{1}{c}{\textbf{MMSU}} \\
\cmidrule(lr){4-7} \cmidrule(lr){8-8} 
& & & Sound$\uparrow$ & Music$\uparrow$ & Speech$\uparrow$ & Average$\uparrow$ & Accuracy $\uparrow$ \\
\midrule
\midrule
\multicolumn{1}{l}
{\uline{\textbf{\textit{Baselines}}}} \\
Qwen2.5-Omni & - & - & 67.87 & 69.16 & 59.76 & 65.60 & 61.32 \\
\midrule
\midrule
\multicolumn{1}{l}{\uline{\textbf{\textit{Ours}}}} \\
SARI & SFT + GRPO + CL & Structured CoT & \textbf{72.75} & 67.22 & \textbf{61.26} & \textbf{67.08} & \textbf{66.01} \\
- & SFT + GRPO + CL & Unstructured CoT & 71.17 & 66.39 & 60.14 & 65.90 & 65.16 \\
\midrule
\bottomrule
\end{tabular}}
\label{tab:qwen2-omni-result}
\end{table*}

Through the aforementioned experiments, we demonstrate that employing curriculum-enhanced GRPO method to instruct Large Audio-Language Models in engaging with structured reasoning about input audio yields significant benefits. As illustrated in Table \ref{tab:mmau_mmsu}, this approach enables our model achieves 16.35\% higher average accuracy than the base model Qwen2-Audio-7B-Instruct on MMAU. Furthermore, our model based on Qwen2.5-Omni achieves 67.08\% SOTA performance on the MMAU test-mini benchmark. It also maintains competitive performance on MMSU, a cross-domain evaluation benchmark, with 66.01\% accuracy, demonstrating robust generalization capabilities. Through the construction of datasets incorporating structured CoT annotations and a two-stage training framework (SFT followed by curriculum-enhanced GRPO), the advantage of our method manifests not only in superior task-specific performance but also in enhanced general reasoning abilities. This proves particularly critical for real-world audio AI assistants, which must reliably process unforeseen queries while delivering human-comprehensible reasoning traces.

\begin{table*}[h!]
\centering
\setlength{\tabcolsep}{4pt} 
\caption{Performance comparison on MMAU Test-mini and MMSU benchmark.\cite{MMAU2024}}
\begin{tabular}{c|ccccc}
\toprule

\multirow{3}{*}{\textbf{Model}} & \multicolumn{4}{c}{\textbf{MMAU Test-mini}} & \multicolumn{1}{c}{\textbf{MMSU}} \\
\cmidrule(lr){2-5} \cmidrule(lr){6-6} 
 & Sound$\uparrow$ & Music$\uparrow$ & Speech$\uparrow$ & Average$\uparrow$ & Accuracy $\uparrow$ \\
 \midrule
\uline{\textbf{\textit{Closed-Source}}} & & & & & \\
GPT4o + Caption & 63.36 & 60.77 & 53.15 & 57.30 & - \\
Gemini Pro 2.0 Flash & 56.46 & 58.68 & 51.65 & 55.60 & - \\
\midrule
\uline{\textbf{\textit{Open-Source}}} & & & & & \\
Qwen2-Audio-7B-Instruct & 54.95 & 50.98 & 42.04 & 49.20 & 35.72 \\
Audio-Reasoner & 60.06 & 64.30 & 60.70 & 61.71 & - \\
R1-AQA & 69.37 & 66.77 & 57.36 & 64.50 & - \\
Qwen2.5-Omni & 67.87 & 69.16 & 59.76 & 65.60 & 61.32 \\
\midrule
\uline{\textbf{\textit{Ours}}} & & & & & \\
\textbf{SARI}( Qwen2-Audio-7B-Instruct based) & 68.55 & 69.01 & 59.09 & 65.55 & 35.36 \\
\textbf{SARI}(Qwen2.5-Omni based) & 72.75 & 67.22 & 61.26 & \textbf{67.08} & \textbf{66.01} \\
\bottomrule
\end{tabular}
\label{tab:mmau_mmsu}
\end{table*}

\section{Conclusion}
We presented a study on improving the reasoning capabilities of audio multimodal models by leveraging explicit chain-of-thought training and curriculum-guided reinforcement learning. We constructed a novel dataset of 32k audio-focused multi-choices questions. Through a series of experiments, we systematically compared explicit reasoning models (that output their thought process) with implicit models (answer-only). Our results show that explicit reasoning models perform better on audio reasoning tasks, confirming findings from prior work that intermediate reasoning helps large models tackle complex problems. We also demonstrated the critical role of curriculum learning in reinforcement fine-tuning. By training the model from easy to hard questions, we achieved higher success rates.  Ultimately, our best model, which employs structured reasoning and curriculum-based GRPO optimization, reached state-of-the-art performance on the challenging MMAU test-mini benchmark and showed superior general reasoning on MMSU, setting a new bar for audio understanding models.

\section{Limitation}
We utilized approximately 32k samples for the entire experiment, yet the dataset remains relatively limited, particularly with a small proportion of speech data. Additionally, our data construction relied entirely on open-source models; we anticipate that employing more advanced models could yield better results. Lastly, our experiments were conducted exclusively with Qwen2-Audio-7B-Instruct and Qwen2.5-Omni, suggesting that the generalizability of our conclusions might require further investigation.


\clearpage
\bibliographystyle{unsrt}
\bibliography{ref}

\appendix

\section{PROMPTS} 
\subsection{Prompts for Reasoning Path Generation} \label{AppendixA}

\begin{tcolorbox}[colframe=blue!60!black, colback=blue!5!white, title=Prompt for Structured Thought Generation (Model: Qwen2.5-72B-Instruct)]

We are now designing a system to generate audio-based chain-of-thought reasoning
data. You will receive an audio clip's textual description, as well as a question and its answer. Your task is to explore in more detail the thinking process from the question to the answer. Your chain of thought will be divided into two main parts, the <think> and <response> sections. In the <think> section, you need to go through four steps: planning, captioning, reasoning, and summarizing. The <think> section is invisible to the user.Therefore, in the <response> section, you need to base on all the reasoning processes and results in the <think> section and provide a final reply based on the question. Finally, your
reply should strictly follow the following format: <THINK><PLANNING> (In this part, analyze the user’s needs and how to complete the task. If the problem is complex, it should be done step by step) </PLANNING><CAPTION> (In this part, conduct an overall analysis of the given audio input content, try to find all the parts related to the question, describe them, and ensure it is completely correct.) </CAPTION><REASONING> (Start reasoning towards
the final answer here) </REASONING><SUMMARY> (Draw appropriate results based on
the reasoning part here) </SUMMARY></THINK><RESPONSE> Give the final answer here
referring to the <THINK> part </RESPONSE> Please strictly follow the format of the sample:

<THINK>

<PLANNING>

The user wants to understand the dynamic changes within the provided audio clip ......
</PLANNING>

<CAPTION>

The audio clip predominantly features static noise. ...... similar to that of a detuned television
or a device failing to receive a signal.

</CAPTION>

<REASONING>
1. Identify changes in Intensity (Volume): The audio’s static noise does not remain at a
constant volume. There are noticeable increases and decreases in loudness throughout the
clip. ...... is struggling to maintain a consistent output, adding to the impression of something
malfunctioning or broken.

</REASONING>

<SUMMARY>

The static noise in the audio is highly dynamic. ...... leading to a sense of disorder and
instability.

</SUMMARY>

</THINK>

<RESPONSE>

The audio presents a static noise, ......The overall effect is one of energetic chaos, preventing
any possibility of calm or predictability.

</RESPONSE>
Note that you have both the question and the answer because it is necessary to ensure the
correctness of the chain of thought. However, in your response, you can only refer to the
content of the question and the audio, and lead to the answer. You must absolutely not assume
that you already know the answer. Please provide a detailed and flexible response with
high-quality logic in both the caption and reasoning sections. If the reasoning part requires
complex logic, you can even propose several different approaches and try them one by one.

Here is the original description: 
*** \{caption\} ***.

The question is: 
*** \{question\} ***

The answer you can refer to: 
*** \{answer\} ***.

Again, don’t mention that you have the answer and the description because they are only here
to help you to design the chain of thought but should not exist in the real-world scenario,
either in the think or response sections.

\vspace{0.2cm}
\end{tcolorbox}

\begin{tcolorbox}[colframe=blue!60!black, colback=blue!5!white, title=Prompt for Unstructured Thought Generation (Model: Qwen2.5-72B-Instruct)]
We are designing a system to generate audio-based chain-of-thought reasoning data. You will receive a textual description of an audio clip, along with a question and its answer. Your task is to simulate a rich, human-like thinking process from the question to the answer.

Your thought process must be written in a single **non-structured paragraph** enclosed within the <THINK> ... </THINK> tags. The content should resemble how a person might naturally think: exploring different possibilities, forming and revising hypotheses, doubting their initial ideas, backtracking, verifying assumptions, and gradually refining their conclusions.

  Strict instructions:
- Do **not** mention or hint at reading, seeing, or receiving any "description," "transcript," or "audio content."
- Do **not** use phrases like "based on the description," "the audio says," "from what I heard," or similar.
- **Only use inferred content directly as if you experienced it firsthand.**
- Avoid any meta-commentary or system-level awareness.

You are encouraged to:
- **Reflect** on your assumptions ("Am I making a leap here?")
- **Backtrack** or change your mind if needed ("Wait, maybe that doesn't make sense...")
- **Validate** parts of your reasoning ("If that were true, what would follow? Does that match the audio description?")
- **Explore multiple interpretations** before settling on one
This process should not look clean or pre-structured — it should flow naturally and feel exploratory and thoughtful. Even though you are given the answer, you must **never** show awareness of it or base your logic on it directly. Use only the audio description and question to build your reasoning.

After the <THINK> section, write a brief final answer in the <ANSWER> section, summarizing your conclusion based on your thought process.

Output Format:

<THINK>

...Your detailed, natural, reflective reasoning goes here...

</THINK>

<ANSWER>

...Your final answer here...

</ANSWER>

Here is the original description:  
*** \{caption\} ***  

The question is:  
*** \{question\} ***  

The answer you can refer to:  
*** \{answer\} ***

Note: The description and answer are only for your internal use to guide the reasoning process. — they must **not** appear or be referenced in the output. And do not assume you know the answer. Your task is to simulate the thinking of someone who is trying to **figure it out from scratch**.
\end{tcolorbox}

\section{Case Studies}
\label{AppendixB}

This section presents response examples from various models derived from Qwen2-Audio-7B-Instruct. The training strategies and CoT paradigms for each model are detailed in Table \ref{tab:qwen2-audio-result}.

\begin{tcolorbox}[colframe=blue!60!black, colback=blue!5!white, title=Model-A]
{\color{blue} 
\textbf{Instruction} \\
Based on the given audio, identify the source of the whoop. Please choose the answer from the following options: ['Human', 'Bird', 'Dog', 'Machine']. \\
Output the final answer in <ANSWER> </ANSWER>.\\} 

<ANSWER> Human</ANSWER>
\end{tcolorbox}

\begin{tcolorbox}[colframe=blue!60!black, colback=blue!5!white, title=Model-B]
{\color{blue} 
\textbf{Instruction} \\
Based on the given audio, identify the source of the whoop. Please choose the answer from the following options: ['Human', 'Bird', 'Dog', 'Machine']. \\
Output the thinking process in <THINK> </THINK>, and divide it into four parts: <PLANNING>, <CAPTION>, <REASONING>, and <SUMMARY>. After that, output the final answer in <ANSWER> </ANSWER>.\\}

<THINK>\\  <PLANNING> Begin thinking</PLANNING>\\  <CAPTION> Thinking about the options</CAPTION>\\  <REASONING> Evaluating Human, Bird, Dog, and Machine</REASONING>\\ <SUMMARY> Drawing a conclusion</SUMMARY>\\</THINK>\\<ANSWER> Human</ANSWER>
\end{tcolorbox}

\begin{tcolorbox}[colframe=blue!60!black, colback=blue!5!white,title=Model-C, float=htbp]
{\color{blue} 
\textbf{Instruction} \\
Based on the given audio, identify the source of the whoop. Please choose the answer from the following options: ['Human', 'Bird', 'Dog', 'Machine']. \\
Output the thinking process in <THINK> </THINK> and final answer in <ANSWER> </ANSWER>.\\} 

<THINK> Analyzing the sound </THINK>\\
<ANSWER> Human</ANSWER>
\end{tcolorbox}

\begin{tcolorbox}[colframe=blue!60!black, colback=blue!5!white, title=Model-D,  breakable]
{\color{blue} 
\textbf{Instruction} \\
Based on the given audio, identify the source of the whoop. Please choose the answer from the following options: ['Human', 'Bird', 'Dog', 'Machine']. \\
Output the thinking process in <THINK> </THINK>, and divide it into four parts: <PLANNING>, <CAPTION>, <REASONING>, and <SUMMARY>. After that, output the final answer in <ANSWER> </ANSWER>.\\}

<THINK>\\
<PLANNING>\\
The user wants to determine the source of the whoop heard in the audio clip. To achieve this, I need to analyze the characteristics of the whoop and compare them to typical sounds produced by humans, birds, dogs, and machines. This involves understanding the nature of the whoop and matching it to the most plausible source.\\
</PLANNING>\\
<CAPTION>\\
The audio clip contains a whoop, which is a loud, sharp vocalization. The whoop is characteristic of human vocalizations and is often associated with excitement or celebration.\\
</CAPTION>\\
<REASONING>\\
1. **Identify the Whoop**: The whoop is a loud, sharp sound that is typically associated with human vocalizations.\\
2. **Compare with Human Whoops**: Human whoops can vary in intensity and frequency but are generally produced by individuals expressing strong emotions such as joy, excitement, or surprise.\\
3. **Compare with Animal Whoops**: Birds can produce chirping or squawking sounds, but these are usually more melodic and less sharp than a human whoop.\\
4. **Compare with Machine Whoops**: Machines can produce various sounds, but these are usually more mechanical and lack the human quality of a whoop.\\
5. **Consider Context**: The whoop is often heard in social gatherings or during celebrations, which further supports the idea that it is a human vocalization.\\
</REASONING>\\
<SUMMARY>\\
Based on the characteristics of the whoop and the typical sounds produced by humans, the most plausible source of the whoop is a human.\\
</SUMMARY>\\
</THINK>\\

<ANSWER>Human</ANSWER>
\end{tcolorbox}

\begin{tcolorbox}[colframe=blue!60!black, colback=blue!5!white, title=Model-E]
{\color{blue} 
\textbf{Instruction} \\
Based on the given audio, identify the source of the whoop. Please choose the answer from the following options: ['Human', 'Bird', 'Dog', 'Machine']. \\
Output the thinking process in <THINK> </THINK> and final answer in <ANSWER> </ANSWER>.\\} 

<THINK>\\Okay, let me think about this. The whoop sounds like it could be coming from a person. It’s definitely a human voice, and it seems to fit the context of a lively, social environment where there’s a lot of background noise from a river and other activities. The whoop isn’t repetitive or mechanical, so it’s unlikely to be a machine. It also doesn’t sound like a bird or a dog making a sound. The combination of the whoop and the ambient sounds makes me lean heavily towards the idea that the source is a human. If it were a dog, it would probably be barking or panting more prominently. A bird wouldn’t make a whoop sound. And a machine isn’t fitting either. So, the most logical choice is the human.\\</THINK>\\

<ANSWER>Human</ANSWER>
\end{tcolorbox}

\begin{tcolorbox}[colframe=blue!60!black, colback=blue!5!white, title=Model-F, breakable]
{\color{blue} 
\textbf{Instruction} \\
Based on the given audio, identify the source of the whoop. Please choose the answer from the following options: ['Human', 'Bird', 'Dog', 'Machine']. \\
Output the thinking process in <THINK> </THINK>, and divide it into four parts: <PLANNING>, <CAPTION>, <REASONING>, and <SUMMARY>. After that, output the final answer in <ANSWER> </ANSWER>.\\}

<THINK>\\
<PLANNING>\\
The user wants to identify the source of the whoop heard in the audio clip. To do this, I need to carefully analyze the audio content, extract relevant details, and reason through the possible sources before arriving at a conclusion.\\</PLANNING>\\<CAPTION>\\The audio clip contains the sound of flowing water and human voices, including whoops. These elements suggest an outdoor setting near a body of water where people are active.\\</CAPTION>\\<REASONING>\\1. **Identify the Sound**:\\   - **Flowing Water**: The audio includes the sound of flowing water, indicating a natural or recreational setting near a body of water.\\   - **Human Voices and Whoops**: The presence of human voices and whoops strongly suggests that the sound is coming from humans.\\2. **Evaluate the Options**:\\   - **Human**: The audio clearly features human voices and whoops.\\   - **Bird**: There are no bird sounds mentioned in the audio description.\\   - **Dog**: No dog sounds are mentioned.\\   - **Machine**: There are no mechanical sounds indicated in the audio description.\\3. **Conclusion Based on Analysis**:\\   - The audio primarily consists of human voices and whoops, making \"Human\" the most likely source of the whoop.\\</REASONING>\\<SUMMARY>\\Based on the analysis of the audio clip, the whoop is most likely coming from a human source.\\</SUMMARY>\\</THINK>\\

<ANSWER>Human</ANSWER>
\end{tcolorbox}

\begin{tcolorbox}[colframe=blue!60!black, colback=blue!5!white, title=Model-G]
{\color{blue} 
\textbf{Instruction} \\
Based on the given audio, identify the source of the whoop. Please choose the answer from the following options: ['Human', 'Bird', 'Dog', 'Machine']. \\
Output the thinking process in <THINK> </THINK> and final answer in <ANSWER> </ANSWER>.\\} 

<THINK>\\Hmm, this whoop sounds pretty human. It’s definitely not a bird, dog, or machine. If it were a machine, it wouldn’t have that kind of emotional or vocal quality. Let’s see if I’m making a leap here. If it were a bird, the whoop would likely be more of a chirp or tweet. A dog might bark or whine, but certainly not whoop. So, if it’s not a machine, it must be a human. That seems to fit best with what I’m hearing. Am I making a leap here? Let’s see if this holds up under scrutiny. If it were a human, wouldn’t the whoop be more of an emotional expression or a social interaction? Yes, that makes sense. So, it’s probably a human whoop.\\</THINK>\\

<ANSWER>Human</ANSWER>
\end{tcolorbox}

\end{document}